\documentclass[letterpaper]{article}
\usepackage[preprint]{aaai2027}
\usepackage[hyphens]{url}
\usepackage{graphicx}
\usepackage{natbib}
\usepackage{caption}
\usepackage{booktabs}
\usepackage{amsmath}
\usepackage{amssymb}
\usepackage{multirow}
\title{Hallucinations Leave a Grounding Signature:\\Verifier-Guided Decoding for Selective Object Correction}
\author{
Lei Yang\textsuperscript{\rm 1,\rm 2}\equalcontrib,
Xinze Liu\textsuperscript{\rm 1,\rm 2}\equalcontrib,
Dayan Wu\textsuperscript{\rm 1}\corresponding,
Ding Wang\textsuperscript{\rm 3},
Hengjie Zhu\textsuperscript{\rm 1,\rm 2},\\
Zihao Zhang\textsuperscript{\rm 1,\rm 2},
Tianzhu Hu\textsuperscript{\rm 1,\rm 2},
Hanqi Wu\textsuperscript{\rm 1,\rm 2},
Peng Fu\textsuperscript{\rm 1},
Zheng Lin\textsuperscript{\rm 1}
}
\affiliations{
\textsuperscript{\rm 1}Institute of Information Engineering, Chinese Academy of Sciences, Beijing, China; wudayan@iie.ac.cn\\
\textsuperscript{\rm 2}School of Cyber Security, University of Chinese Academy of Sciences, Beijing, China\\
\textsuperscript{\rm 3}Department of Applied Mathematics and Statistics, Johns Hopkins University, Baltimore, MD, USA
}

\begin{document}
\maketitle

\begin{abstract}
Large vision-language models (LVLMs) often hallucinate objects that are absent from an image. Despite recent progress, existing mitigation methods still lack reliable object-level grounding diagnostics and therefore tend to apply coarse-grained interventions, which can impair visual understanding, shorten responses, and reduce coverage of genuinely grounded objects. The key challenge is thus to detect, during generation, whether each emerging object mention is supported by reliable visual evidence, so that hallucination can be mitigated selectively. Yet output confidence reflects next-token plausibility rather than visual support, allowing language priors to make absent objects appear certain. We show that the missing diagnostic evidence is encoded in an Intrinsic Grounding Signature (IGS), a distributed signed attention pattern that remains informative for such confident hallucinations. Based on IGS, we propose Verifier-Guided Decoding (VGD), a decoding framework in which a lightweight verifier examines each emerging object mention, rolls back the KV cache when the mention is identified as high risk, suppresses the object and its synonyms, and regenerates the affected continuation. Because VGD intervenes only on object mentions identified as high risk, it reduces object hallucination while preserving the model's original visual understanding and grounded object coverage. Experiments on CHAIR and AMBER-G show that VGD achieves state-of-the-art object hallucination reduction: at @rec90, it cuts AMBER-G CHAIR by 43.6\% while retaining 99.6\% of grounded-object coverage, and reduces CHAIR-MSCOCO CHAIR$_i$/CHAIR$_s$ by 37.0\%/30.4\% without shortening captions.
\end{abstract}

\section{Introduction}
Large vision-language models (LVLMs) have advanced rapidly through strong visual encoders~\cite{radford2021learning}, cross-modal alignment~\cite{li2023blip2,alayrac2022flamingo}, and large-scale instruction tuning~\cite{liu2023visual,wang2024qwen2,bai2025qwen25vl}. Yet LVLMs still hallucinate objects, fluently and confidently describing entities absent from the image~\cite{rohrbach2018object,li2023evaluating}. Such errors are particularly concerning in safety-critical settings, where fluent natural language can mask visually unsupported claims. This risk is further amplified in longer descriptions, as unsupported object mentions can accumulate even when next-token confidence remains high. In this work, we focus on object hallucination, commonly measured by CHAIR~\cite{rohrbach2018object} as the fraction of object mentions or sentences lacking visual support.

\begin{figure}[t]
\centering
\includegraphics[width=\columnwidth]{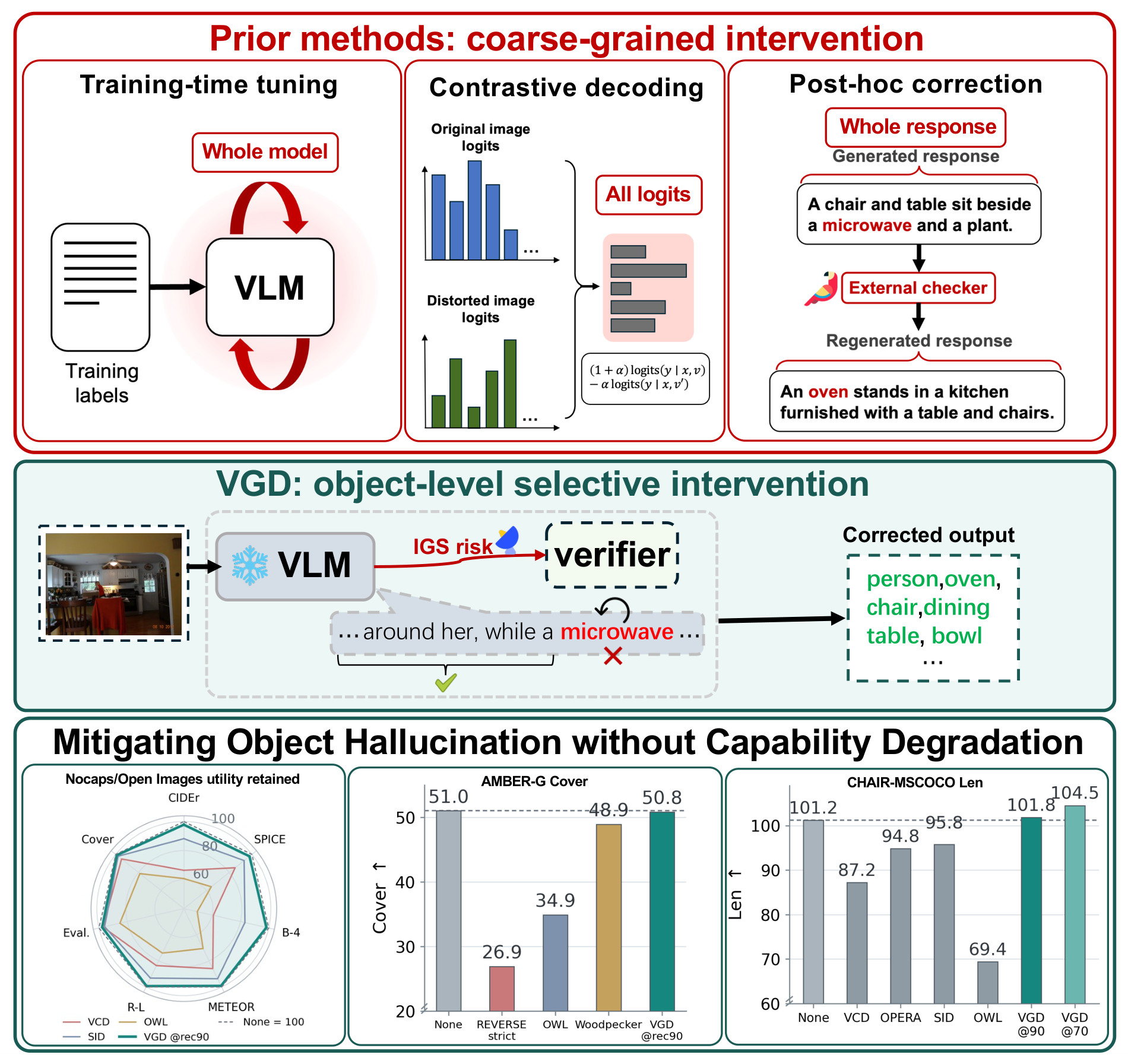}
\caption{Intervention scope and retained utility. Prior paradigms act at model, distribution, or response level; VGD verifies each emerging mention and locally rewrites only high-risk ones. Bottom: at @rec90, cross-dataset utility is normalized to None $=100$, while Cover and Len retain their original scales.}
\label{fig:motivation}
\end{figure}

Existing remedies intervene at one of the three stages: training-time methods reshape model behavior using additional supervision, inference-time methods steer token distributions or attention patterns during decoding, and post-hoc methods revise completed responses~\cite{liu2024robust,gunjal2024detecting,leng2024mitigating,huang2024opera,yin2024woodpecker,zhou2024analyzing}. Although these approaches address hallucination from different angles, they generally operate at the model, decoding-trajectory, or full-response level, rather than making calibrated grounding decisions for individual object mentions as they emerge. Figure~\ref{fig:motivation} shows that this broad scope may lower hallucination at the cost of shorter responses or reduced object coverage, degrading overall generation quality. In contrast, VGD performs mention-level verification during generation and rewrites only object mentions identified as high-risk hallucinations.

The shared missing capability is object-level risk assignment before intervention: the key question is not how aggressively generation should be suppressed, but where correction is actually warranted. Because object mentions constitute discrete factual commitments about the visual input, correction should target unsupported mentions without disturbing grounded content or the surrounding accepted context. Achieving this requires an object-conditioned signal that is available as each mention begins to form during decoding. Output confidence is insufficient because it reflects contextual next-token preference rather than visual grounding, allowing faithful and hallucinated mentions to appear equally certain. This presents a central challenge: identifying reliable internal evidence of insufficient visual support during generation.

\begin{figure}[t]
\centering
\includegraphics[width=\columnwidth]{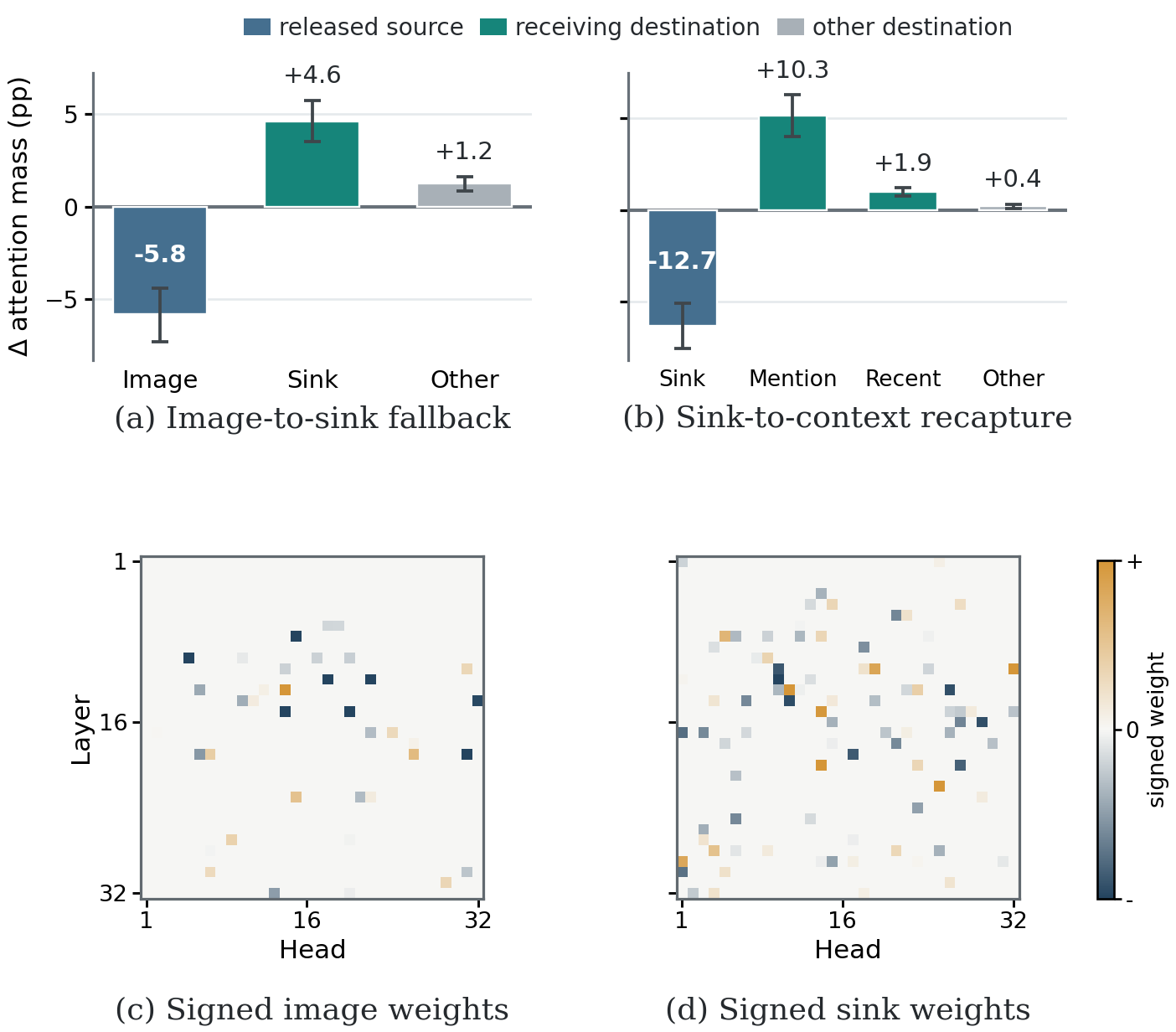}
\caption{Causal and structural evidence for signed IGS. Evidence removal induces visual-to-sink fallback (a) or mention/history capture (b), while sparse signed weights span heads and layers (c,d). Bars show paired target-mask-minus-sham changes with $95\%$ intervals over $180$ interventions.}
\label{fig:mechanism}
\end{figure}

To find such evidence, we probe a frozen LVLM during token-by-token generation. As illustrated in Figure~\ref{fig:mechanism}(a,b), visual ungrounding produces a structured, head-dependent redistribution rather than a uniformly weaker attention signal. In visual-grounding heads, missing support weakens object-to-image query--key alignment, causing released attention to fall back to the stable BOS sink and increase sink mass. In linguistic-continuation heads, autoregressive decoding must preserve a coherent continuation, so the emerging mention and recent-context keys capture more query attention, drawing mass away from BOS and decreasing sink mass. Controlled evidence removal confirms these complementary responses, motivating a signed cross-head representation rather than scalar attention averaging.

Because these complementary responses have opposite signs, scalar image- or sink-attention averages can cancel rather than reveal the diagnostic evidence. We call their distributed, object-conditioned pattern the \emph{Intrinsic Grounding Signature} (IGS). By preserving the direction of head-specific attention shifts, IGS exposes whether an emerging object mention lacks visual support, turning object hallucination from a post-hoc output error into a generation-time event that can be localized and intercepted.

Building on this capability, we propose Verifier-Guided Decoding (VGD), a lightweight framework that couples correction scope to object-level grounding risk. The verifier reads each emerging mention's IGS and accepts low-risk content unchanged. When risk exceeds a threshold, VGD rolls the key-value cache back to the mention start, suppresses the object and its synonyms, and regenerates only the affected span. This risk-conditioned procedure preserves the frozen LVLM and accepted prefix, while a single threshold controls intervention aggressiveness and exposes the hallucination--coverage trade-off.

Experiments across architectures and benchmarks show that IGS provides reliable object-level grounding diagnosis and that VGD substantially reduces object hallucination while largely preserving grounded-object coverage and caption length at the high-recall operating point. Its controllable threshold exposes rather than conceals the cost of stricter correction, while cross-dataset transfer without verifier refitting supports the generality of the learned grounding signal. Together, these results establish model-internal grounding diagnosis as a practical basis for selective hallucination mitigation without broadly suppressing the model's generation capability.

Our contributions are three-fold:
\begin{itemize}
    \item We establish generation-time object grounding diagnosis to overcome coarse-grained mitigation, enabling selective correction without degrading visual understanding.
    \item We uncover IGS, which enables object hallucinations to be directly localized during generation, and build VGD to selectively rewrite only high-risk mentions while preserving the frozen LVLM.
    \item Across six architectures and three benchmarks, IGS reaches $0.927$--$0.939$ AUROC; VGD achieves state-of-the-art reduction with near-base Cover and Len, no-refit transfer, and $1.61\times$ greedy cost.
\end{itemize}

\begin{figure*}[t]
\centering
\includegraphics[width=\textwidth]{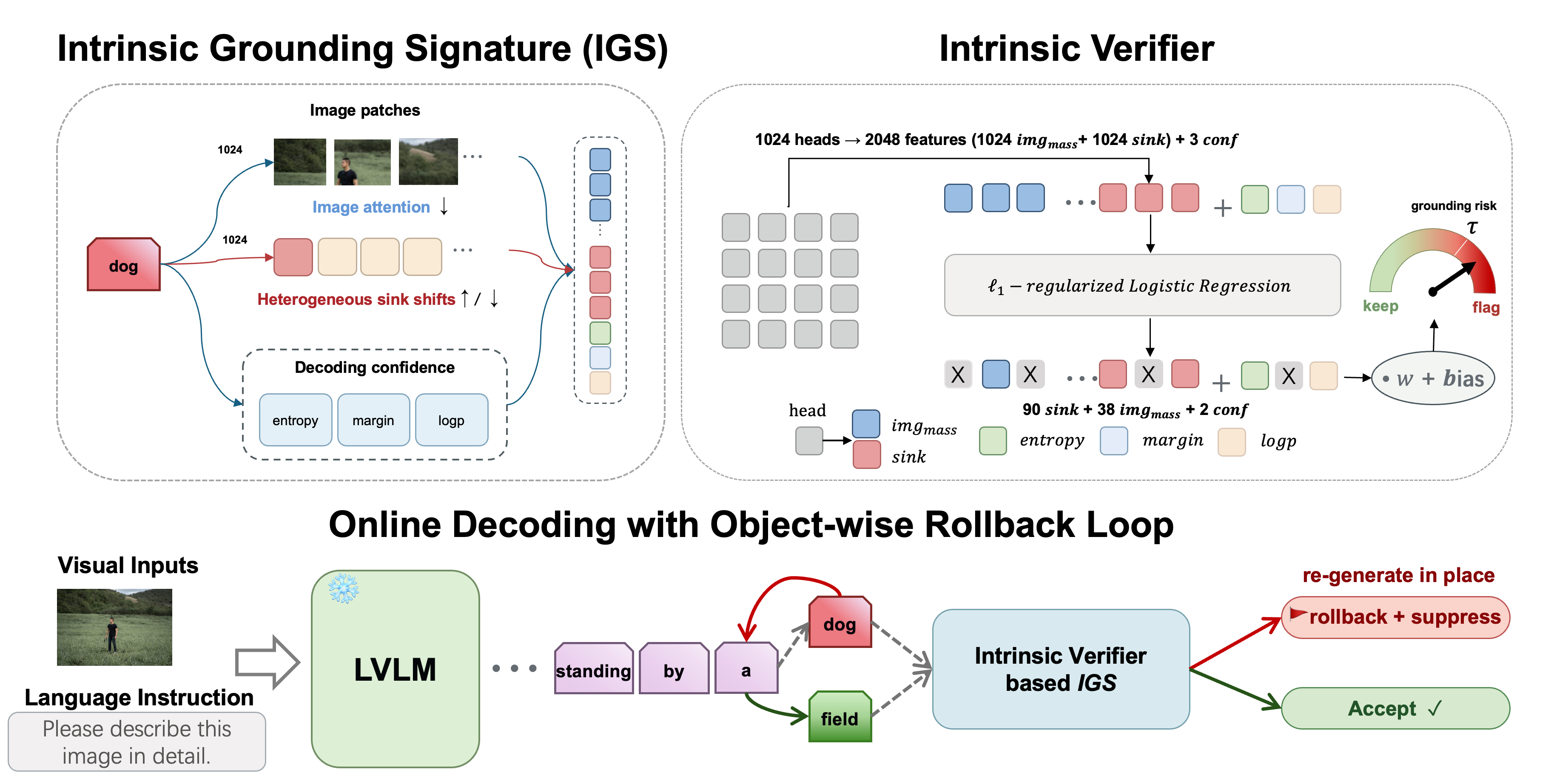}
\caption{From IGS to verifier-guided decoding. Top: distributed image/sink attention forms per-object IGS, which an offline $\ell_1$-sparse signed readout maps to grounding risk. Bottom: low-risk mentions are accepted, whereas high-risk ones trigger KV-cache rollback, synonym suppression, and local regeneration with the accepted prefix preserved.}
\label{fig:method}
\end{figure*}

\section{Related Work}

\textbf{Hallucination mitigation.}
Training-time alignment modifies LVLMs using additional supervision~\cite{liu2024robust,yu2024rlhfv,gunjal2024detecting}, while external systems inspect and revise completed responses~\cite{yin2024woodpecker,zhou2024analyzing}. Inference-time methods contrast images or layers~\cite{leng2024mitigating,chuang2024dola}, assemble global and local evidence~\cite{an2025agla}, or redirect visual, history, and sink attention~\cite{liu2024paying,huang2024opera,huo2025sid,yu2025owl,shukla2026sage,zhang2026savaa}. Other approaches align attention with information flow~\cite{zhao2025aligning} or steer internal representations after detecting visual--linguistic conflict~\cite{liu2026vli}. These methods differ in mechanism, but generally intervene on model parameters, decoding distributions, internal states, or completed responses rather than using a calibrated grounding decision to bound correction at each object span.

\textbf{Grounding signals and selective verification.}
Internal-state studies probe attention ratios, intermediate attention, or hidden representations~\cite{chuang2024lookback,jiang2024devils,kogilathota2026halp}. Object-level detectors further score global--local similarity~\cite{park2025glsim}, preceding-token attention~\cite{hoang2026pas}, or Bayesian posteriors over decoding and description signals~\cite{zohrabi2026haloprobe}, while REVERSE learns hallucination-aware retrospective resampling~\cite{wu2025reverse}. Similar attention magnitudes for faithful and hallucinated objects motivate structured signals~\cite{wang2026same}. Sequence-start sinks and visual attention sinks provide related routing evidence~\cite{clark2019bert,voita2019analyzing,xiao2024efficient,sun2024massive,kang2025visualsink,wang2025mirage}, but absolute or averaged attention can erase opposing head responses. What remains missing is a span-aligned, actionable grounding signal. IGS instead recovers signed image--sink redistribution across heads, and VGD converts this span-aligned diagnosis into prefix-preserving local rollback.
\section{Intrinsic Grounding Diagnosis}
\label{sec:igs}

Before designing a decoding intervention, we ask what information inside a frozen LVLM can determine whether an emerging object mention is visually supported. This section moves from candidate scalar cues to IGS, then tests why its image and sink components respond in different directions.

\subsection{Problem Formulation and Grounding Signals}
Given an image $I$ and a textual instruction, a visual encoder and projector map $I$ to $N$ image tokens $\tilde V=[\tilde v_1,\ldots,\tilde v_N]$. Together with textual prefix $U$ and the generated history, they form decoder input $X$. At step $t$, hidden state $h_t$ induces $p(x_t\mid I,X_{<t})=\operatorname{softmax}(W_o h_t)$, where $x_t$ is the next token, $X_{<t}$ its preceding input, and $W_o$ the output projection. For object-token position $o$, let $\alpha_{o,j}^{(l,h)}$ be causal attention from head $h\in\{1,\ldots,H\}$ in layer $l\in\{1,\ldots,L\}$ to key position $j$, i.e., let $\alpha_{o,j}^{(l,h)}$ denote the attention weight from query position $o$ to key position $j$. We retain image attention mass $g$, the total attention to image-token positions $\mathcal{I}_{\mathrm{img}}$, and sink attention $s$, the attention assigned to the sequence-start sink:
\begin{equation}\label{eq:grounding_signals}
\begin{aligned}
g^{(l,h)}(o)&=\sum_{j\in\mathcal{I}_{\mathrm{img}}}\alpha_{o,j}^{(l,h)},
\qquad
s^{(l,h)}(o)=\alpha_{o,j_{\mathrm{sink}}}^{(l,h)}.
\end{aligned}
\end{equation}
We also retain entropy, top-two margin, and log probability as a confidence vector $c(o)\in\mathbb{R}^3$. Positions are one-based. For LLaVA-1.5, $j_{\mathrm{sink}}=1$ corresponds to the BOS token \texttt{<s>} and corresponds to key index $0$ in code; architecture-specific mappings are provided in the supplement. We use CHAIR to assign object-level grounded/hallucinated labels to recognized object spans in greedy captions and collect $g^{(l,h)}$, $s^{(l,h)}$, and $c$ from more than four thousand LLaVA-generated object mentions.

\subsection{From Confident Visual Ungrounding to Signed IGS}
Selective correction requires evidence that identifies which emerging object mention lacks visual support, rather than suppressing generation globally. We first test whether output confidence provides that evidence.

\paragraph{Confidence fails as a grounding cue.}
Entropy, top-two margin, and generated-token log probability are standard confidence and uncertainty cues~\cite{hendrycks2017baseline,guo2017calibration,malinin2021uncertainty}, yet jointly reach only $0.70$--$0.73$ AUROC with strongly overlapping grounded and hallucinated distributions. High output confidence therefore does not imply visual grounding. We call this failure mode \textbf{confident visual ungrounding}: a confident object commitment without sufficient visual support.

\paragraph{Visual evidence weakens without becoming uncertainty.}
Hallucinated mentions reduce the all-head image-mass average in Equation~\eqref{eq:image_mass_average},
\begin{equation}\label{eq:image_mass_average}
\bar g(o)=\frac{1}{LH}\sum_{l,h}g^{(l,h)}(o)
\end{equation}
with a Cohen's $d$ of about $1.4$, while this scalar reaches $0.81$ AUROC. Thus, even when local attention peaks appear comparable~\cite{wang2026same}, aggregate visual support weakens without reliable output uncertainty. This scalar remains insufficient because Figure~\ref{fig:mechanism}(c,d) shows sparse image and sink coefficients dispersed across heads, layers, and signs. Some heads raise sink attention while others lower it, so unsigned averaging cancels structure and isolated probes discard complementary routes. The supplement reports the full distributions and confirms that the separation is not explained by caption position or category frequency.

We call this object-conditioned pattern the \emph{Intrinsic Grounding Signature} (IGS), which is read from attention already produced by the frozen LVLM and calibrated at the span level in Section~\ref{sec:verifier}. Position and category controls jointly reach only $0.794$ AUROC, whereas signed IGS remains at $0.935$ after including them. Within matched category and mention-position strata, it retains $0.882$ AUROC. Full distributions and controls are in the supplement.

\subsection{Causal Anatomy of Signed IGS}
Generating an object mention couples two internal routes. Visual-grounding heads locate object evidence, whereas linguistic-continuation heads maintain the forming mention and its local context. We test whether the opposite sink signs expose a handoff between these routes by masking target evidence for $180$ grounded mentions spanning all $80$ COCO categories and comparing it with an area-matched sham mask under the same caption. We exclude final evaluation images, form pairs within each image, and cross-fit heads before held-out intervention analysis, while collection and bootstrap details are provided in the supplement.

\paragraph{Visual fallback and linguistic compensation.}
In visual-grounding head L31H16, target masking breaks the match between the object and its image evidence, so image attention falls from $0.264$ to $0.206$. BOS remains a globally available, image-invariant anchor, so the released mass flows primarily to the sink, which rises from $0.488$ to $0.534$ in Figure~\ref{fig:mechanism}(a). In linguistic-continuation head L15H26, the visual target disappears but the emerging lexical hypothesis and recent context remain. Because autoregressive prediction must still continue coherently, these surviving language cues capture the query as an implicit compensation. Consequently, current-mention and recent-text mass rise by $0.103$ and $0.019$, while sink attention falls by $0.127$ and image attention remains unchanged in Figure~\ref{fig:mechanism}(b). The model can therefore continue fluently and confidently after visual grounding has failed. Both heads retain the expected verifier sign in four of five cross-fitted splits, showing that these responses are not isolated examples.

\paragraph{Why the signs differ.}
After omitting the $l,h$ superscripts for one fixed head, let $q_o$ be the query at $o$, $k_j$ the key at position $j$, and $d_h$ their dimension. Define $z_j=q_o^\top k_j/\sqrt{d_h}$, $s=j_{\mathrm{sink}}$, and $Z_o=\operatorname{LSE}_j z_j=\log\sum_j e^{z_j}$, so $\alpha_{o,s}=e^{z_s-Z_o}$. With $\Delta$ denoting target-mask minus sham, the exact identity is
\begin{equation}
\Delta\log\alpha_{o,s}=\Delta z_s-\Delta Z_o.
\label{eq:sink_decomposition}
\end{equation}
BOS precedes all image tokens, so masking leaves its key and value fixed. Sink-up is denominator-driven: visual-key competition collapses while the BOS score changes little, lowering $Z_o$ and releasing probability to the default anchor. Sink-down is query-driven: learned local-language routing strengthens current-mention and recent-text matches while weakening the generic BOS match. Cross-fitted family and value-path analyses in the supplement confirm that both responses recur and reach the head output. They reveal two sides of one \emph{grounding-to-language handoff}: the visual route withdraws while the language route sustains the unsupported mention. Signed IGS is necessary because unsigned averaging erases this complementary evidence.

\section{Verifier-Guided Decoding}
\label{sec:method}

Figure~\ref{fig:method} connects the three stages of VGD: per-object IGS construction, a sparse grounding-risk readout fitted offline, and online selective correction. During generation, low-risk mentions are accepted unchanged, while only high-risk mentions trigger KV-cache rollback, synonym suppression, and local regeneration. The base LVLM and the accepted prefix remain intact throughout.

\subsection{Intrinsic Grounding Verifier}
\label{sec:verifier}
As summarized at the top of Figure~\ref{fig:method}, the verifier maps per-object IGS to scalar visual-ungrounding risk through a sparse signed readout.

\paragraph{Feature construction.}
For an object-token position $o$, the verifier instantiates IGS with two attention feature groups: image-mass $g^{(l,h)}(o)$ and sink attention $s^{(l,h)}(o)$ over all layers and heads. Together with the confidence vector $c(o)$ defined above, the feature vector is
\begin{equation}\label{eq:phi}
\begin{gathered}
\phi(o)=
\big[\{g^{(l,h)}(o)\}_{l,h},\{s^{(l,h)}(o)\}_{l,h},c(o)\big],\\
\phi(o)\in\mathbb{R}^{2LH+3}.
\end{gathered}
\end{equation}
For LLaVA-1.5-7B, $\phi(o)$ has $2\times32\times32+3=2051$ dimensions. For a multi-token mention, image and sink features are averaged over all tokens in its matched span, while $c(o)$ is evaluated when predicting the first object token, and $o$ henceforth denotes this span-level representation.

\paragraph{Sparse linear readout.}
We use an $\ell_1$-regularized logistic regression:
\begin{align}
r(o)&=\sigma\!\big(w^\top\phi(o)+b\big), \label{eq:risk}\\
(\hat w,\hat b)&=\arg\min_{w,b}\frac1n\sum_{i=1}^{n}
\mathcal{L}_{\mathrm{BCE}}\big(y_i,r(o_i)\big)+\lambda\lVert w\rVert_1.
\label{eq:risk_train}
\end{align}
Here $\sigma$ is the logistic sigmoid, with weights $w\in\mathbb{R}^{2LH+3}$ and intercept $b\in\mathbb{R}$. Among $n$ calibration mentions, $o_i$ is mention $i$, $y_i\in\{0,1\}$ equals $1$ when ungrounded, and $\mathcal{L}_{\mathrm{BCE}}$ is binary cross-entropy, while hats denote fitted minimizers and $\lambda$ weights the $\ell_1$ penalty. Thus, $r(o)$ is predicted ungrounding risk. Signed coefficients align opposing sink responses, while sparsity selects a compact subset from thousands of dimensions. Each backbone fits this model once on automatically CHAIR-labeled mentions from calibration images disjoint from final evaluation, and the same data selects $\tau$ by target grounded-object recall without test access. Because the LVLM remains frozen and no auxiliary VLM is used, online scoring requires one dot product. Full solver and split settings are in the supplement.

\subsection{Verifier-Guided Local Decoding}
\label{sec:local_decoding}
The bottom of Figure~\ref{fig:method} summarizes the resulting object-level rollback loop. During greedy decoding, a lightweight noun recognizer prioritizes multi-word objects and then matches single-word mentions. Once the complete surface form is available, verification uses the position $p_o$ of its first subtoken and the span aggregation from Section~\ref{sec:verifier}. VGD accepts the mention when $r(o)\leq\tau$ and otherwise performs three local operations. First, it crops the KV cache to the prefix before $p_o$, restoring the preceding state without recomputation. Second, it forms suppression set $\mathcal{B}=\{o\}\cup\mathrm{Syn}(o)$, where $\mathrm{Syn}(o)$ is the synonym set of $o$, and masks the corresponding word logits. Third, greedy decoding resumes at $p_o$ to rewrite the entire affected continuation. The base parameters and all prefix states before the mention remain unchanged.

The single threshold $\tau$ controls intervention frequency because high thresholds retain more grounded detail, whereas lower thresholds flag more mentions and reduce hallucination at a greater false-positive cost. VGD therefore exposes a continuous trade-off between coverage and hallucination rather than committing to one hidden operating point.

\begin{table*}[t]
\centering
\small
\setlength{\tabcolsep}{4.0pt}
\renewcommand{\arraystretch}{1.0}
\begin{tabular}{@{}lllccccccc@{}}
\toprule
\textbf{Base LVLM} & \textbf{Method Type} & \textbf{Method}
& \multicolumn{3}{c}{\textbf{CHAIR-MSCOCO}}
& \multicolumn{4}{c}{\textbf{AMBER-G}} \\
\cmidrule(lr){4-6}\cmidrule(lr){7-10}
& & & CHAIR$_i(\downarrow)$ & CHAIR$_s(\downarrow)$ & Len
& CHAIR $(\downarrow)$ & Cover $(\uparrow)$ & Hall $(\downarrow)$ & Cog $(\downarrow)$ \\
\midrule
\multirow{16}{*}{LLaVA-1.5-7B}
& None & None & 15.4 & 50.0 & 101.2 & 7.8 & 51.0 & 36.4 & 4.2 \\
& \multirow{7}{*}{Gen-Adjust}
& VCD$^{\dagger\ddagger}$ & 14.9 & 48.6 & 87.2 & 8.7 & 50.8 & 36.6 & 3.4 \\
& & OPERA$^{\dagger\ddagger}$ & 14.6 & 49.6 & 94.8 & 6.4 & 48.6 & 27.3 & 2.7 \\
& & DoLa & 14.1 & 51.6 & -- & 7.6 & 51.6 & 36.0 & 4.0 \\
& & SID$^{\dagger\ddagger}$ & 13.9 & 48.2 & 95.8 & 5.4 & 52.6 & 28.3 & 2.5 \\
& & OWL$^{\dagger\ddagger}$ & 9.2 & 26.2 & 69.4 & 4.4 & 34.9 & 14.1 & 0.8 \\
& & AGLA & 14.1 & 43.0 & -- & -- & -- & -- & -- \\
& & MEMVR & 13.0 & 46.6 & -- & -- & -- & -- & -- \\
& \multirow{5}{*}{w/ Train}
& EOS & 12.3 & 40.2 & -- & 5.1 & 49.1 & 22.7 & 2.0 \\
& & HALVA & 11.7 & 41.4 & -- & 6.6 & \textbf{53.0} & 32.2 & 3.4 \\
& & HA-DPO & 11.0 & 38.2 & -- & 6.7 & 49.8 & 30.9 & 3.3 \\
& & REVERSE$_{(\tau=0.003)}$ & 10.3 & 37.0 & -- & 6.0 & 52.2 & 30.4 & 3.0 \\
& & REVERSE$_{(\tau=0.0003)}$ & 6.1 & 13.6 & -- & 4.0 & 26.9 & 10.2 & 0.9 \\
& Post-hoc Refine
& Woodpecker & 14.8 & 45.8 & -- & 6.9 & 48.9 & 30.4 & 3.6 \\
& \multirow{2}{*}{Verify (ours)}
& \textbf{VGD @rec90} & 9.7 & 34.8 & 101.8 & 4.4 & 50.8 & 19.1 & 1.4 \\
& & \textbf{VGD @rec70} & \textbf{4.6} & \textbf{13.4} & 104.5 & \textbf{2.8} & 38.2 & \textbf{8.2} & \textbf{0.6} \\
\midrule
\multirow{3}{*}{Qwen2-VL-7B}
& None & None & 8.6 & 27.4 & 86.4 & 7.4 & \textbf{68.4} & 44.3 & 4.4 \\
& \multirow{2}{*}{Verify (ours)}
& \textbf{VGD @rec90} & 4.7 & 15.8 & 86.1 & 1.9 & 55.6 & 10.2 & 0.6 \\
& & \textbf{VGD @rec70} & \textbf{3.2} & \textbf{10.0} & 86.5 & \textbf{1.8} & 40.4 & \textbf{6.5} & \textbf{0.4} \\
\midrule
\multirow{3}{*}{Shikra-7B}
& None & None & 16.5 & 53.4 & 102.2 & 11.4 & \textbf{51.1} & 47.9 & 5.5 \\
& \multirow{2}{*}{Verify (ours)}
& \textbf{VGD @rec90} & 6.8 & 20.2 & 101.9 & 4.3 & 47.4 & 17.6 & 1.2 \\
& & \textbf{VGD @rec70} & \textbf{4.1} & \textbf{10.4} & 103.0 & \textbf{3.4} & 37.3 & \textbf{10.9} & \textbf{0.7} \\
\bottomrule
\end{tabular}
\caption{Unified CHAIR-MSCOCO and AMBER-G results. @rec90 emphasizes coverage and @rec70 suppression; Len is mean CHAIR word count. $\dagger$ and $\ddagger$ mark CHAIR-MSCOCO and AMBER-G results reproduced under the unified setting, respectively. Unmarked baseline results are taken from the same unified evaluation reported by REVERSE~\cite{wu2025reverse}.}
\label{tab:main}
\end{table*}

\section{Experiments}
\label{sec:experiments}
We evaluate grounding, diagnostic evidence, cross-dataset transfer, ablations, and online cost.

\subsection{Setup}
\textbf{Models and benchmarks.}
We evaluate IGS detection across six LLaVA, Qwen, InstructBLIP, and Shikra configurations and online decoding on LLaVA-1.5-7B, Qwen2-VL, and Shikra. CHAIR-MSCOCO reports CHAIR$_i$/CHAIR$_s$/Len. AMBER-G~\cite{wang2023amber} reports grounded-object coverage (Cover), the percentage of responses with any hallucination (Hall), and the recall of contextually associated but absent targets (Cog). Lower is better except for Cover. On all $4{,}500$ NoCaps validation images, reference metrics assess caption quality and verified Open Images labels support OI-VN after excluding unknown pairs~\cite{agrawal2019nocaps,kuznetsova2020openimages}. Because prompts differ, Len is compared only within each benchmark.

\textbf{Baselines and protocol.}
Local Table~\ref{tab:main} runs share backbone, prompt, parser, and split; their $512/1024$-token caps are nonbinding because all outputs end before $512$. Unmarked rows come from REVERSE. Methods span generation adjustment~\cite{leng2024mitigating,huang2024opera,chuang2024dola,huo2025sid,yu2025owl}, training~\cite{wu2025reverse}, and post-hoc correction~\cite{yin2024woodpecker}; Cover and Len expose conservative generation.

\textbf{Implementation details.}
None and VGD decode greedily; other local baselines use their method-specific decoders. Each frozen backbone fits a class-balanced $\ell_1$ verifier on image-disjoint CHAIR mentions. We denote by @rec$q$ the calibration-only threshold that retains $q\%$ of grounded mentions, and evaluate @rec98 through @rec70 without test access. NoCaps reuses the fixed LLaVA verifier fitted from $4{,}031$ nonoverlapping mentions, and rollback is capped at $40$. Solver and split details are in the supplement.

\subsection{Selective Mitigation and Controllable Frontier}

\paragraph{Primary operating point.}
VGD's primary @rec90 point reduces hallucination without the conservative-output shortcut. Table~\ref{tab:main} shows that LLaVA AMBER-G CHAIR falls from $7.8$ to $4.4$ while Cover remains $50.8$ versus $51.0$. CHAIR$_i$/CHAIR$_s$ likewise fall from $15.4/50.0$ to $9.7/34.8$, while Len remains $101.8$ versus $101.2$. At the same AMBER-G CHAIR, OWL retains only $34.9$ Cover with a mean length of $69.4$ words, whereas VGD occupies the higher-coverage, full-length region and exceeds SID and higher-coverage REVERSE at comparable coverage.
Mention- and sentence-level CHAIR decrease together, so the gain is not confined to a few repeatedly failing captions. As summarized in Figure~\ref{fig:motivation}, stable Cover and Len show that @rec90 preserves effective object understanding rather than lowering CHAIR by describing less of the image.

\paragraph{\mbox{Stricter correction.}}
For LLaVA, VGD at @rec70 reaches AMBER-G CHAIR $2.8$ while trading Cover to $38.2$, still above strict REVERSE at $26.9$. The same @rec90 mechanism lowers AMBER-G CHAIR from $7.4$ to $1.9$ on Qwen2-VL and from $11.4$ to $4.3$ on Shikra with essentially unchanged CHAIR caption length. Model-dependent Cover shifts motivate explicit operating points, while consistent hallucination reduction supports architectural recurrence; a representative case is in Supplementary Fig.~A4.

\begin{figure}[t]
\centering
\includegraphics[width=\columnwidth]{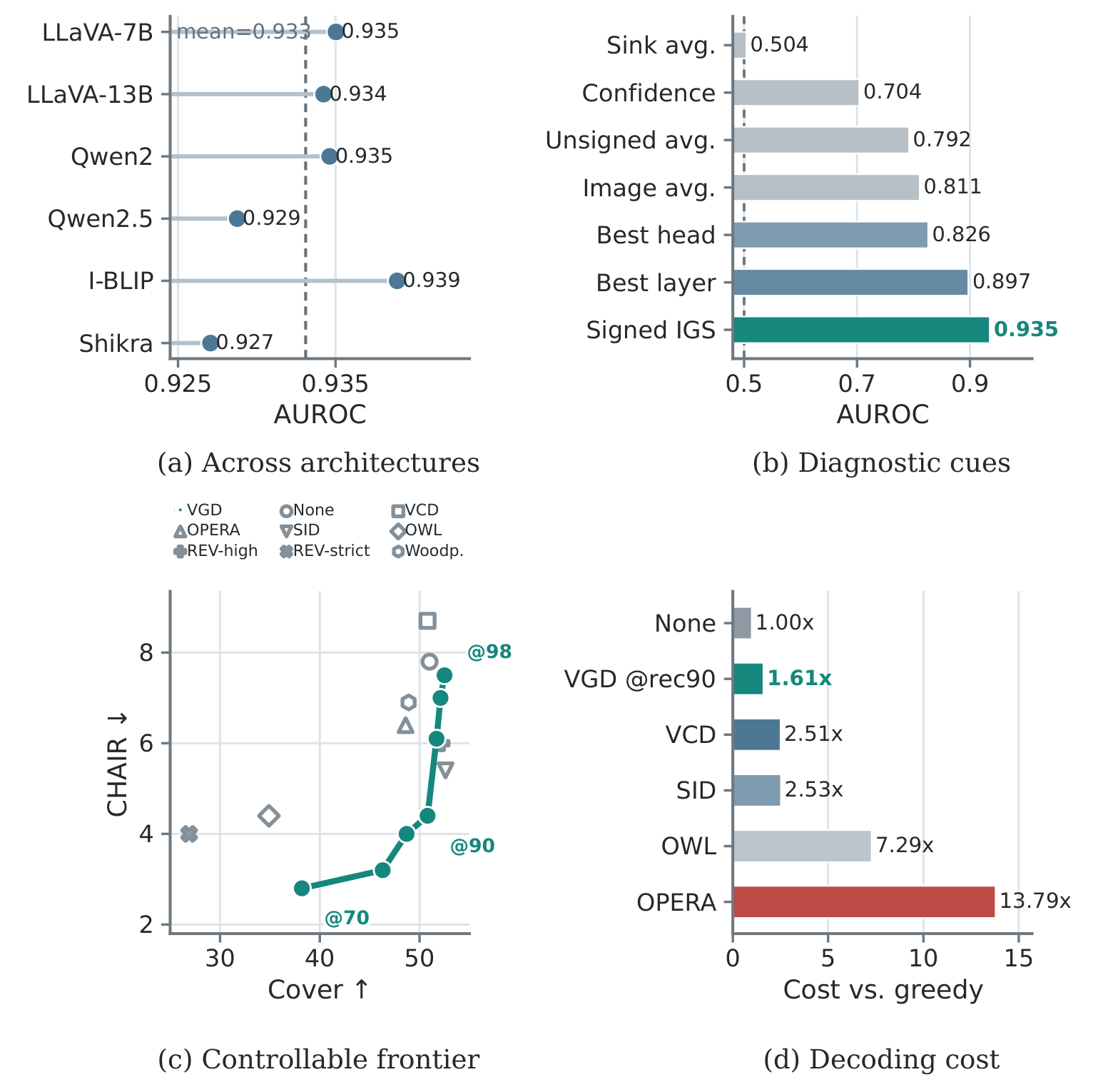}
\caption{IGS diagnosis, operating points, and efficiency. (a) Cross-architecture recurrence. (b) Signed versus isolated cues. (c) AMBER-G Cover and CHAIR frontier. (d) Matched relative wall-clock cost.}
\label{fig:evidence}
\end{figure}

\paragraph{One threshold, an explicit frontier.}
Figure~\ref{fig:evidence}(c) sweeps CHAIR/Cover from $7.5/52.5$ at @rec98 to $2.8/38.2$ at @rec70. Thus, @rec90 is the balanced high-coverage point, whereas @rec70 is stricter. Figure~\ref{fig:evidence}(d) reports the matched decoding cost of the primary @rec90 point. Because all points share one verifier, the curve exposes how intervention frequency trades hallucination against coverage.

\begin{table*}[!t]
\centering
\small
\setlength{\tabcolsep}{2.8pt}
\begin{tabular}{@{}lccccc@{\hspace{5pt}}cccccc@{}}
\toprule
& \multicolumn{5}{c}{NoCaps caption quality}
& \multicolumn{6}{c}{Open Images object-existence evaluation} \\
\cmidrule(lr){2-6}\cmidrule(lr){7-12}
Method & CIDEr$\uparrow$ & SPICE$\uparrow$ & B-4$\uparrow$ & METEOR$\uparrow$ & R-L$\uparrow$
& OI-VN$_s\downarrow$ & OI-VN$_i\downarrow$ & Eval.$\uparrow$ & Cover$\uparrow$ & Len & Rollback \\
\midrule
None & \textbf{103.85} & \textbf{16.49} & \textbf{39.66} & \textbf{30.99} & \textbf{59.59}
& 8.90 & 6.28 & \textbf{84.36} & \textbf{20.00} & 12.40 & -- \\
 VCD& 69.00& 14.05& 24.13& 26.62& 49.83& 10.22& 7.33& 81.96& 18.98& 13.72&--\\
 SID& 91.62& 15.36& 32.99& 29.02& 55.51& 8.75& 6.35& 81.00& 19.59& 12.12&--\\
 OWL& 63.04& 10.59& 19.62& 21.87& 44.11& 7.45& 5.73& 71.89& 15.75& 7.95&--\\
VGD @rec90 & 101.64 & 16.14 & 39.02 & 30.72 & 59.17
& 7.89 & 5.60 & 82.49 & 19.83 & 12.44 & 0.32 \\
VGD @rec70 & 86.39 & 13.95 & 34.66 & 28.88 & 56.49
& \textbf{5.04} & \textbf{3.92} & 67.89 & 15.46 & 12.83 & 1.62 \\
\bottomrule
\end{tabular}
\caption{Unified NoCaps/Open Images transfer. OI-VN follows the official hierarchy and excludes unknowns; Eval. and Len are evaluable-caption rate and mean words.}
\label{tab:nocaps}
\end{table*}

\subsection{Evaluating IGS Diagnosis}
\paragraph{Recurrence across architectures.}
IGS recurs even when its architecture-specific readout changes. Because each backbone fits an independent verifier, Figure~\ref{fig:evidence}(a) does not transfer one coefficient vector but instead tests whether the same grounding state remains recoverable. AUROC stays within $0.927$--$0.939$ across six LVLM configurations, with a mean of $0.933$.

\paragraph{\mbox{Why signed structure matters.}}
Signed cue aggregation, rather than any isolated cue, supplies the diagnostic gain. Figure~\ref{fig:evidence}(b) shows only $0.704$ AUROC from confidence, while the best layer reaches $0.897$ and unsigned aggregation reaches $0.792$, both remaining below signed IGS at $0.935$. Simpler sink and image averages reach $0.504$ and $0.811$, and even the best single head reaches only $0.826$. Moreover, hallucinations raise sink attention in $455$ heads but lower it in $569$, matching the complementary routes in Figure~\ref{fig:mechanism}. Thus, IGS recovers aligned evidence from heterogeneous directions rather than selecting one favorable head, layer, or high-magnitude statistic, with position and category controls provided in the supplement.

\subsection{Cross-Dataset Generalization}
\label{sec:cross_dataset}

Table~\ref{tab:nocaps} tests transfer on all $4{,}500$ NoCaps validation images under matched LLaVA decoding~\cite{agrawal2019nocaps}. VGD reuses the fixed CHAIR verifier and thresholds without refitting or image overlap. OI-VN applies the official Open Images hierarchy to verified labels and excludes unknown pairs~\cite{kuznetsova2020openimages}. OI-VN$_s$ is the percentage of evaluable captions with a verified-negative mention, while OI-VN$_i$ is the percentage of known mentioned classes that are verified negative. No image-level Open Images labels or reference captions are used during decoding.

At @rec90, VGD lowers OI-VN$_s$/OI-VN$_i$ from $8.90/6.28$ to $7.89/5.60$. Both paired $95\%$ confidence intervals exclude zero, with McNemar $p=6.1\!\times\!10^{-9}$, while Cover/Len and CIDEr/SPICE remain near base. VCD worsens OI-VN to $10.22/7.33$, SID reduces CIDEr to $91.62$, and OWL lowers OI-VN only while reducing Cover/Len to $15.75/7.95$ and producing $174$ empty captions. The normalized utility profile in Figure~\ref{fig:motivation} summarizes this contrast across seven quality and coverage measures. @rec70 exposes the cost of stricter suppression. The ranking persists when hierarchy propagation is disabled and only directly verified labels are scored, with full statistics in the supplement. Together, lower verified-negative error from a fixed CHAIR verifier and near-base caption quality support transfer of grounding diagnosis rather than dataset-specific refitting.

\subsection{Ablations and Efficiency}
\begin{table}[!t]
\centering
\footnotesize
\setlength{\tabcolsep}{1.0pt}
\begin{tabular}{@{}lccccc@{}}
\toprule
Components & AUROC$\uparrow$ & CHAIR$\downarrow$ & Cover$\uparrow$ & Hall$\downarrow$ & Cog$\downarrow$ \\
\midrule
LLaVA-1.5-7B & -- & 7.8 & 51.0 & 36.4 & 4.2 \\
\hspace{0.6em}+ Confidence features & 0.702 & 8.5 & \textbf{51.3} & 35.8 & 4.1 \\
\hspace{1.4em}+ Sink mass & 0.887 & 6.3 & 47.7 & 25.5 & 2.5 \\
\hspace{2.2em}+ Image mass & 0.892 & 4.7 & 47.0 & 20.2 & 1.6 \\
\hspace{3.0em}+ $\ell_1$ regularization & \textbf{0.935} & \textbf{4.4} & 50.8 & \textbf{19.1} & \textbf{1.4} \\
\bottomrule
\end{tabular}
\caption{Cumulative verifier and decoding ablations at @rec90. All preceding verifier rows use L2 logistic regression. The final row replaces L2 with L1 over the same complete feature set, improving intervention precision.}
\label{tab:readout_ablation}
\end{table}

\paragraph{Diagnostic quality and intervention utility.}
With rollback and @rec90 fixed, Table~\ref{tab:readout_ablation} isolates trigger quality. Confidence alone worsens CHAIR. Sink and image evidence reduce CHAIR/Hall/Cog to $6.3/25.5/2.5$ and $4.7/20.2/1.6$, yet Cover remains below base. The sparse signed readout reaches $0.935$ AUROC and $4.4$ CHAIR while restoring Cover from $47.0$ to $50.8$, a $3.8$-point gain. Thus, aggregate AUROC alone does not determine online utility: threshold-local false positives govern unnecessary rollback, while the retained prefix shapes rewrite quality. A false positive removes a grounded span, whereas a true positive helps only if regeneration yields a faithful alternative. With recall and rollback fixed, L1 reflects a cleaner decision boundary rather than a stronger intervention.

\paragraph{Intervention frequency and computational cost.}
Across $4{,}500$ NoCaps images, @rec90 activates in $661$ captions, averages $0.322$ rollbacks, and reaches the $40$-retry cap in only $13$ cases; $85.3\%$ remain on the greedy path and none is empty. At @rec70, $2{,}447$ captions activate, averaging $1.618$ rollbacks, with $68$ capped. Capped rates of $0.29\%$ and $1.51\%$ show that the cap is a safeguard, not the source of improvement. Figure~\ref{fig:evidence}(d) measures $2.73$s per caption versus $1.69$s for greedy decoding, a $1.61\times$ cost below the compared interventions. All methods use one H20 and identical image ordering, excluding loading time. Thus, @rec90 leaves most generations untouched, whereas @rec70 pays for stronger suppression through more frequent local rewriting rather than empty or failed outputs. This concentration of work on a minority of captions explains why verifier precision, rather than a higher retry ceiling, is the main route to a better frontier. Timing and retry summaries are in the supplement.

\FloatBarrier
\section{Conclusion}
Object hallucination mitigation requires deciding where correction is warranted rather than broadly suppressing generation. We identify IGS, a distributed signed attention pattern that exposes object-level visual ungrounding as a mention forms and explains why scalar cues can miss unsupported objects. VGD converts this internal evidence into per-object risk, locally rolling back and regenerating only high-risk spans while leaving the base LVLM unchanged. Across architectures and benchmarks, VGD achieves a stronger hallucination--coverage frontier without shortening descriptions at the high-recall point, while transfer without verifier refitting supports the cross-dataset reuse of IGS. These results turn object hallucination from a post-hoc error into a generation-time event that can be localized and selectively corrected.

\bibliography{references}

\end{document}